%% file: icml2021.tex
\documentclass{article}

\usepackage{hyperref}

\usepackage[accepted]{icml2021}

\icmltitlerunning{An End-to-End Framework for Molecular Conformation Generation via Bilevel Programming}

\usepackage{microtype}
\usepackage{graphicx}
\usepackage{subfigure}
\usepackage{booktabs} 
\usepackage{multirow}
\usepackage{amssymb}

\input{math_commands}

\usepackage{threeparttable}
\usepackage{tablefootnote}
\usepackage{algorithm, algorithmic}
\usepackage{xspace}

\def\ie{{\it i.e.}}

\def\eg{{\it e.g.}}

\def\diff{{\operatorname{d}}}

\def\Jian#1 {\textcolor{red}{[Jian: #1]}}

\newcommand{\modelname}{ConfVAE\xspace}

\begin{document}

\twocolumn[
\icmltitle{An End-to-End Framework for Molecular Conformation Generation \\via Bilevel Programming}


\icmlsetsymbol{equal}{*}

\begin{icmlauthorlist}
\icmlauthor{Minkai Xu}{mila,udem}
\icmlauthor{Wujie Wang}{mit}
\icmlauthor{Shitong Luo}{pku}
\icmlauthor{Chence Shi}{mila,udem}\\
\icmlauthor{Yoshua Bengio}{mila,udem,cifar}
\icmlauthor{Rafael Gomez-Bombarelli}{mit}
\icmlauthor{Jian Tang}{mila,cifar,hec}
\end{icmlauthorlist}

\icmlaffiliation{mila}{Mila - Qu\'ebec AI Institute, Canada}
\icmlaffiliation{udem}{Universit\'e de Montr\'eal, Canada}
\icmlaffiliation{mit}{Massachusetts Institute of Technology, USA}
\icmlaffiliation{pku}{Peking University, China}
\icmlaffiliation{cifar}{Canadian Institute for Advanced Research (CIFAR), Canada}
\icmlaffiliation{hec}{HEC Montr\'eal, Canada}

\icmlcorrespondingauthor{Minkai Xu}{minkai.xu@umontreal.ca}

\icmlkeywords{Machine Learning, ICML}

\vskip 0.3in
]



\printAffiliationsAndNotice{\icmlEqualContribution} 

\input{00-abstract}
\input{10-intro}
\input{20-background}
\input{30-method}
\input{31-model}
\input{32-training}
\input{40-exp}
\input{41-conf-gen}
\input{42-distance}
\input{50-related}
\input{60-conclusion}

\input{65-acknowledgment}

\bibliography{icml2021}
\bibliographystyle{icml2021}

\onecolumn
\newpage
\twocolumn
\input{70-app}

\end{document}

%% file: math_commands.tex
\usepackage{amsmath,amsfonts,bm}

\def\eqref#1{equation~\ref{#1}}

\def\1{\bm{1}}

\def\vzero{{\bm{0}}}

\def\vd{{\bm{d}}}

\def\vr{{\bm{r}}}

\def\mI{{\bm{I}}}

\def\mR{{\bm{R}}}

\DeclareMathAlphabet{\mathsfit}{\encodingdefault}{\sfdefault}{m}{sl}
\SetMathAlphabet{\mathsfit}{bold}{\encodingdefault}{\sfdefault}{bx}{n}

\def\gE{{\mathcal{E}}}

\def\gG{{\mathcal{G}}}

\def\gL{{\mathcal{L}}}

\def\gN{{\mathcal{N}}}

\def\gV{{\mathcal{V}}}

\def\sR{{\mathbb{R}}}
\def\sS{{\mathbb{S}}}



%% file: 00-abstract.tex
\begin{abstract}

Predicting molecular conformations (or 3D structures) from molecular graphs is a fundamental problem in many applications. Most existing approaches are usually divided into two steps by first predicting the distances between atoms and then generating a 3D structure through optimizing a distance geometry problem. However, the distances predicted with such two-stage approaches may not be able to consistently preserve the geometry of local atomic neighborhoods, making the generated structures unsatisfying. In this paper, we propose an end-to-end solution for molecular conformation prediction called \modelname based on the conditional variational autoencoder framework. Specifically, the molecular graph is first encoded in a latent space, and then the 3D structures are generated by solving a principled bilevel optimization program. Extensive experiments on several benchmark data sets prove the effectiveness of our proposed approach over existing state-of-the-art approaches. Code is available at \url{https://github.com/MinkaiXu/ConfVAE-ICML21}.

\end{abstract}

%% file: 10-intro.tex
\section{Introduction}



Recently we have witnessed much success of deep learning for molecule modeling in a variety of applications 
ranging from molecule property prediction~\citep{gilmer2017neural} and molecule generation~\citep{you2018graph,shi2020graphaf} to retrosynthesis planning~\cite{shi2020graph}. In these applications, molecules are generally represented as graphs with atoms as nodes and covalent chemical bonds as edges. Although this is empirically effective, in reality molecules are better represented as 3D structures (also known as \textit{conformations}), where each atom is characterized by 3D Cartesian coordinates. Such 3D structures are also more intrinsic and informative, determining many chemical and biological properties such as chemical sensing and therapeutic interactions with proteins. 


However, determining the 3D structures from experiments is challenging and costly. Effectively predicting valid and low-energy conformations has been a very important and active topic in computational chemistry. Traditional computational approaches are typically based on Markov chain Monte Carlo (MCMC) or molecular dynamics (MD)~\citep{de2016role} to propose conformations combined with simulations to assign energies through cheap empirical potentials or expensive quantum chemical simulations~\citep{ballard2015exploiting}.
Recently, there is growing interest in developing machine learning approaches~\cite{mansimov19molecular, simm2019generative,xu2021learning} to model the conditional distribution $p(\mR|\gG)$ of stable conformations $\mR$ given the molecular graph $\gG$ by training on a collection of molecules with available stable conformations. Specifically, two recent works~\citep{simm2019generative,xu2021learning} propose to first predict the distances between atoms and then generate molecular conformations based on the predicted distances by solving a distance geometry problem~\citep{liberti2014euclidean}.
Such approaches based on distance geometry effectively take into account the rotation and translation invariance of molecular conformations and have hence achieved very promising performance.

However, there is still a significant limitation for these two-stage approaches, which predict the distances and conformations separately: the predicted distances might not be able to properly preserve the 3D geometry of local atomic neighborhoods. Some invalid combinations of distances could be assigned a high likelihood according to the distance prediction model. The errors in these distances could be significantly exaggerated by the distance geometry program of the second stage, yielding unrealistic outlier samples of 3D structures. This is not surprising as the distance prediction model is trained by maximizing the factorized likelihood of distances while our end goal is to predict valid and stable conformations. We propose to effectively address this issue with an end-to-end solution which directly predicts the conformation given the molecular graph. Indeed, in a related problem of predicting  3D structures of proteins (\textit{a.k.a.} protein structure prediction) based on amino-acid sequences, the recent huge success of the AlphaFold2 algorithm shows the importance and effectiveness of developing an end-to-end solution compared to the previous AlphaFold algorithm (though exact details of AlphaFold2 algorithm are still lacking)~\cite{senior2020improved,jumper2020high}.

In this paper, we propose such an end-to-end solution called \modelname for molecular conformation generation, based on bilevel programming. To model the rotational and translational invariance of conformations, we still take the pairwise distances among atoms as intermediate variables. However, instead of learning to predict distances by minimizing errors in the space of distance, we formulate the whole problem as bilevel programming~\cite{franceschi2018bilevel}, with the distance prediction problem and  the distance geometry problem for conformation generation being simultaneously optimized. The whole framework is built on the conditional variational autoencoder  (VAE) framework~\cite{kingma2013auto}, in which the molecular graph is first encoded into the VAE latent space, and the conformations are generated based on the latent variable and molecular graph. During training, we iteratively sample a set of distances from the distance prediction model, generate the 3D structures by minimizing an inner objective (defined by the distance geometry problem), and then update the distance prediction model by optimizing the outer objective, \ie, the likelihood directly defined on the conformations.

To the best of our knowledge, \modelname is the first method for molecular conformation generation which can be trained in an end-to-end fashion and at the same time keep the property of rotational and translational invariance. Extensive experiments demonstrate the superior performance of the proposed method over existing state-of-the-art approaches on several widely used benchmarks including conformation generation and distance distribution modeling. We also verify that the end-to-end objective is of vital importance for generating realistic and meaningful conformations.

%% file: 20-background.tex
\section{Background}

\subsection{Problem Definition}

\textbf{Notations.} Following existing work~\citep{simm2019generative,xu2021learning}, each molecule is represented as an attributed atom-bond graph $\gG = \langle \gV, \gE \rangle$, where $\gV$ is the set of vertices representing atoms and $\gE$ is the set of edges representing inter-atomic bonds. Each node $v$ in $\gV$ describes the chosen atomic features such as element type. Each edge $e_{uv}$ in $\gE$ describes the corresponding chemical bond connecting $u$ and $v$, and is labeled with its bond type. Since the distances of bonds existing in the molecular graph are not sufficient to determine an unique conformation (\eg due to so-called \emph{internal rotations} around the axis of the bond), we adopt the common pre-processing methodology in existing works~\cite{simm2019generative,xu2021learning} to expand the graphs by incorporating \textit{auxiliary} edges, which force multi-hop distance constraint eliminating some ambiguities in the 3D conformation, as elaborated in Appendix~\ref{app:sec:preprocess}.

For the geometry $\mR$, each atom in $\gV$ is represented by a 3D coordinate vector $\vr \in \sR^{3}$, and the full set of positions $\{ \vr_v \}_{v \in \gV}$ is represented by the matrix $\mR \in \sR^{|\gV| \times 3}$. Let $d_{uv}$ denote the Euclidean distance $\| \vr_u - \vr_v\|_2$ between the $u^{th}$ and $v^{th}$ atom, then all the distances between connected nodes $\{ d_{uv }\}_{e_{uv} \in \gE}$ can be summarized as a vector $\vd \in \sR^{|\gE|}$. 

\textbf{Problem Definition.} The problem of \emph{molecular conformation generation} is a conditional generation process, where the goal is to model the conditional distribution of molecular conformations $\mR$ given the graph $\gG$, \ie, $p(\mR | \gG)$. 

\subsection{Bilevel Optimization} \label{sec:bk:hpo}
	
Bilevel programs are defined as optimization problems where a set of variables involved in the (outer) objective function are obtained by solving another (inner) optimization problem~\cite{colson2007overview}. Formally, given the outer objective function $F$ and the inner objective $H$, and the corresponding outer and inner variables $\theta$ and $w$, a bilevel program can be formulated by
\begin{equation}
\label{eq:bprog}
    \min_{{\theta}} F(w_{\theta}) \ \ \text{such that} \ \  w_{\theta} \in \arg \min_w H(w, \theta).
\end{equation}
Bilevel programs have shown effectiveness in a variety of situations such as hyperparameter optimization, adversarial and multi-task learning, as well as meta-learning~\cite{maclaurin2015gradient, bengio2000gradient,bennett2006model, flamary2014learning, munoz2017towards, franceschi2018bilevel}.

Typically solving \eqref{eq:bprog} is intractable since the solution sets of $w_\theta$ may not be available in closed form~\citep{bengio2000gradient}. A common approach is to replace the exact minimizer of the inner object $H$ with an approximation solution, which can be obtained through an iterative optimization dynamics $\Phi$ such as stochastic gradient descent (SGD)~\cite{domke2012generic, maclaurin2015gradient, franceschi2017forward}. Starting from the initial parameter $w_0$, we can get the approximate solution $w_{\theta, T}$ by running $T$ iterations of the inner optimization dynamics $\Phi$, \ie, $w_{\theta, T}=\Phi(w_{\theta, T-1}, \theta)=\Phi(\Phi(w_{\theta, T-2}, \theta), \theta)$, and so on. In the general case where $\theta$ and $w$ are real-valued and the objectives and optimization dynamics is smooth, the gradient of the object $F(w_{\theta, T})$ \textit{w.r.t.} $\theta$, named \emph{hypergradient} $\nabla_{\theta} F(w_{\theta, T})$, can be computed by:
\begin{equation}
\label{eq:hypergrad}
      \nabla_{\theta} F(w_{\theta, T}) = \partial_{w} F(w_{\theta,T}) \nabla_{\theta} w_{\theta, T} 
\end{equation}
where $\partial$ denotes the partial derivative to compute the Jacobian on immediate variables
while $\nabla$ denotes a total derivative taking into account the recursive calls to $F$. The above gradient can be efficiently calculated 
by unrolling the optimization dynamics with back-propagation, \ie, reverse-mode automatic differentiation \cite{griewank2008evaluating}, where we repeatedly substitute $w_{\Phi, t} = \Phi(w_{\theta, t-1}, \theta)$ and apply the chain rule. 


%% file: 30-method.tex
\section{Implicit Distance Geometry}

In this section we elaborate on the proposed end-to-end framework. We first present a high-level description of our bilevel formulation in Sec.~\ref{subsec:overview-bilevel}. Then we present the model schematic and training objectives in Sec.~\ref{subsec:model}. Finally we show how to learn the model via hypergradient descent in Sec.~\ref{subsec:model-learning} and how to draw samples in Sec.~\ref{subsec:model-sampling}.

\subsection{Overview}
\label{subsec:overview-bilevel}


Since a molecule can have multiple stable conformations, we model the distribution of conformations $\mR$ conditioning on molecular graph $\gG$ (i.e. $p(\mR|\gG)$) with a conditional variational autoencoder (CVAE)~\citep{kingma2013auto}, in which a latent variable $z$ is introduced to model the uncertainty in molecule conformation generation. 
The CVAE model includes  a prior distribution of latent variable $p_\psi(z|\gG)$ and a decoder $p_\theta(\mR|z,\gG)$ to capture the conditional distribution of $\mR$ given $z$. During training, we also involve an additional inference model (encoder) $q_\phi(z|\mR,\gG)$. The encoder and decoder are jointly trained to maximize the evidence lower bound (ELBO) of the data log-likelihood:
\begin{equation}
\label{eq:elbo}
    \begin{aligned}
    \log P(\mR|\gG) \geq & \mathbb{E}_{z \sim q_{\phi}(z | \mR, \mathcal{G})}\left[\log p_{\theta}(\mR | z, \mathcal{G})\right] \\
    &-D_{\mathrm{KL}}\left[q_{\phi}(z | \mR, \mathcal{G}) \| p_{\psi}(z | \mathcal{G})\right] \\
    \triangleq &  \gL_{ELBO}(\theta,\phi,\psi),
    \end{aligned}
\end{equation}
The ELBO can be interpreted as the sum of the negative reconstruction error $\gL_{recon}$ (the first term) and a latent space prior regularizer
$\gL_{prior}$ (the second term). In practice, $q_{\phi}(z | \mR, \mathcal{G})$ and $p_{\psi}(z | \mathcal{G})$ are all modeled as diagonal Gaussians $N (z|\mu_\phi(\mR, \mathcal{G}), \sigma_\phi(\mR, \mathcal{G}))$ and  $N (z|\mu_\psi(\gG), \sigma_\psi(\gG))$, whose mean and standard deviation are predicted by graph neural networks. To efficiently optimize the ELBO during training, sampling from $q_{\phi}(z | \mR, \mathcal{G})$ is done by reparametrizing $z$ as $z_\phi = \mu_\phi(\mR, \mathcal{G}) + \sigma_\phi(\mR, \mathcal{G}) \cdot \epsilon$, where $\epsilon \sim \gN(\vzero, \mI)$.

With similar encoder and prior models, the key differences among different methods lie in the architecture and learning method of the decoder (generator) model $p_{\theta}(\mR | z, \mathcal{G})$, \ie, how to parameterize the decoder and train it with respect to the reconstruction loss $\gL_{recon}$. Let $D_\theta(z, \gG)$ denote the decoder function 
taking prior $z$ and graph $\gG$ to obtain a
distance vector, we now elaborate how we formulate the optimization problem of the decoder as a bilevel program:

\textbf{Inner objective:} Directly generating conformations as Cartesian coordinates heavily depends on the arbitrary rotation and translation. Therefore, previous effective approaches~\cite{simm2019generative,xu2021learning} instead make the decoder generate inter-atomic distances $\vd$, \ie, $\vd_{\theta,\phi} = D_\theta(z_\phi, \gG)$. The distances $\vd$ are taken as intermediate variables to generate conformations, which are invariant to rotation and translation.
To generate a conformation $\mR$, one needs to first generate the set of distances $\vd$, and then post-process $\vd$ to obtain the 3D positions $\mR$, by solving a distance geometry optimization problem:
\begin{equation}
\label{eq:implicitdg}
\begin{aligned}
    \mR_{\theta,\phi} & = \arg\min_\mR H(\mR, D_\theta(z_\phi,\gG))\\
    & = \arg\min_\mR H(\mR, \vd_{\theta,\phi})\\
    & = \arg\min_\mR \Big\{ \sum_{e_{uv} \in \gE} \big( \| \vr_u  - \vr_v \|_2 - d_{uv} \big)^2 \Big\},
\end{aligned}
\end{equation}
which we take as the inner loop objective. 

\textbf{Outer objective:} Ultimately, we
are interested in directly minimizing the generalization error on 3D structures to make the generated conformation consistent with the ground-truth up to rotation and translation. 
The post-alignment Root-Mean-Square Deviation (RMSD) is a widely used metric for this purpose. To calculate this metric, another conformation $\hat{\mR}$ is first obtained by an alignment function $\hat{\mR} = A(\mR, \mR^*)$, which rotates and translates the reference conformation $\mR^*$ to have the smallest distance to the generated one $\mR$ according to the RMSD metric:
\begin{equation}
\label{eq:rmsd}
    \operatorname{RMSD}(\mR, \hat{\mR})=\Big(\frac{1}{n} \sum_{i=1}^{n}\|\mR_{i}-\hat{\mR_{i}}\|^{2}\Big)^\frac{1}{2}.
\end{equation}
where $n$ is the number of atoms. Then the reconstruction objective $\gL_{recon}$ can be written as:
\begin{equation}
\begin{aligned}
\label{eq:rmsd-outer}
    F(\mR_{\theta,\phi}) & = \log p_{\theta}(\mR | z, \mathcal{G})\\
    &=-\sum_{i=1}^{n} \sum_{j=1}^{3}\left(\mR_{ij}-A(\mR, \mR^*)_{ij}\right)^{2},
\end{aligned}
\end{equation}
which is the outer loop objective for computing the reconstruction loss and maximize the log-likelihood.

\begin{figure*}[!t]
    \centering
    \includegraphics[width=0.99\textwidth]{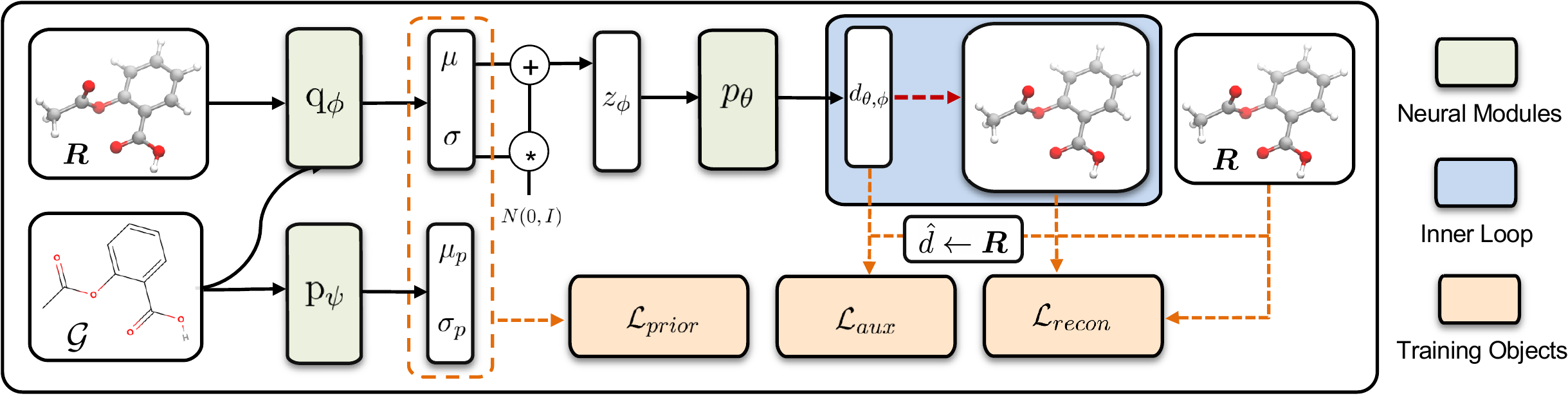}
    \caption{The overall framework of \modelname. At training time, given the graph $\gG$ and conformation $\mR$, we: 1) compute the distributions of $q_\phi(z|\gG, \mR)$ and $p_\psi(z|\gG)$, and calculate $\gL_{prior}$; 2) sample $z$ from $q_\phi$ by reparameterization, and feed it into the decoder (generator) $p_\theta$ to generate inter-atomic distances $\vd$, after which we can obtain an auxiliary objective $\gL_{aux}$ from the true distances $\hat{\vd}$ derived from $\mR$; 3) run the inner loop (distance geometry) to recover the 3D structure from $\vd$, and compute the reconstruction RMSD loss $\gL_{recon}$. The model is trained end-to-end by optimizing the sum of three object components $\gL_{prior}$, $\gL_{aux}$ and $\gL_{recon}$.}
    \label{fig:model}
\end{figure*}

\textbf{Bilevel program:} Now we can consider \eqref{eq:implicitdg} and \eqref{eq:rmsd-outer} as the inner and outer objectives of a bilevel programming problem. In this formulation, the outer objective aims to model the true conditional distribution $p(\mR|\gG)$, and the inner objective solves  for the conformation given a set of predicted distances. By taking the expectation over latent variable $z$, the resulting bilevel program for calculating the reconstruction term $\gL_{recon}$in \eqref{eq:elbo} can be written as:
\begin{align}
    \label{eq:outer-loop}
    & \max_{\theta,\phi} \mathbb{E}_{z \sim q_{\phi}(z | \mR, \mathcal{G})}\left[F(\mR_{\theta,\phi}, \theta)\right]\\
    \label{eq:inner-loop}
    \text{such }&\text{that } \mR_{\theta,\phi} = \arg\min_\mR H(\mR, D_\theta(z_\phi,\gG)).
\end{align}
The derived bilevel problem is still challenging because: 1) the solution of conformation structure in the inner problem is not available in closed form; 2) computing this expectation exactly over the continuous latent space is intractable. Thus, in practice we compute an empirical estimation of the output with a variational inference model and the reparametrization trick.
We elaborate on how we address these issues in the following parts.

%% file: 31-model.tex
\subsection{Generative Model}
\label{subsec:model}

\begin{figure*}[!t]
    \centering
    \includegraphics[width=0.99\textwidth]{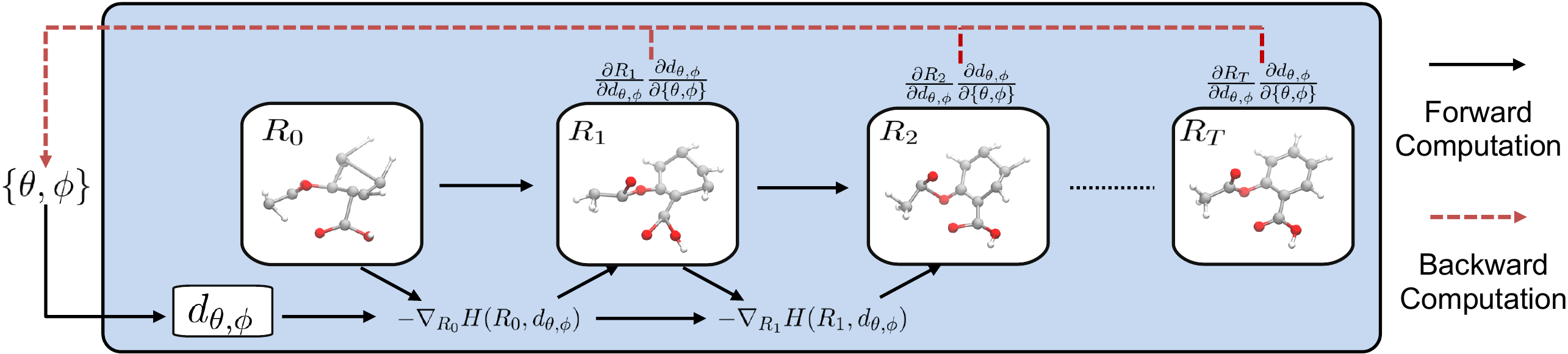}
    \caption{Schematic illustration of the forward and backward computational graph through the inner loop (distance geometry optimization). We repeatedly update $\mR$ with the gradient $\nabla_\mR H$ during the forward computation, and accumulate hypergradients $\nabla_{\theta,\phi}\mR$ to update parameters $\theta$ and $\phi$ from backward computation.}
    \label{fig:inner}
\end{figure*}


We now have the tools needed to define our conditional generative model of molecular conformation.
The cornerstone of all modules (encoder, prior and decoder) is message-passing neural networks (MPNNs)~\cite{gilmer2017neural}, which is a variant of graph neural networks that achieves state-of-the-art performance in representation
learning for molecules~\cite{scarselli2008graph,bruna2013spectral,duvenaud2015convolutional,kipf2016semi,kearnes2016molecular,schutt2017schnet}. The MPNN directly operates on the graph representation $\gG$ and is invariant to graph isomorphism. In each convolutional (message passing) layer, atomic embeddings are updated by aggregating the information from neighboring nodes.

For the encoder $q_\phi(z|\mR,\gG)$ and prior $p_\psi(z|\gG)$, we use the same MPNN architecture as  \citet{mansimov19molecular,simm2019generative}. Since bilevel optimization has a relatively high memory cost, we use an ordinary differential equation (ODE)-based continuous normalizing flow~\cite{chen2018neural} (CNF) for the decoder $p_\theta(\mR|z,\gG)$, which has constant memory cost. We describe the details of our decoder model below.

\textbf{Decoder Architecture.} As illustrated in Sec.~\ref{subsec:overview-bilevel}, our decoder is composed of two cascaded levels: a distance prediction model $D_\theta(z, \gG)$ that decodes $z$ back into a set of distances $\vd$, and a differentiable distance geometry procedure to recover geometry $\mR$ from distances $\vd$. The model $D_\theta(z, \gG)$ is implemented as a conditional extension of the CNF which transforms noise variables $\vd(t_0)$ (also the initial distances in the CNF ODE trajectory) sampled from the prior distribution $\gN(\vzero, \mI)$ to final distances $\vd=\vd(t_1)$. The transformation is conditioned on the latent variable $z$ as well as the graph $\gG$:
\begin{equation}
\begin{aligned}
\label{eq:cgf}
    \vd &= D_\theta(z, \gG) \\
    &= \vd(t_0) + \int_{t_0}^{t_1} g_\theta(\vd(t), t , \gG, z) \diff t, 
\end{aligned}
\end{equation}
where $g_\theta$ is an MPNN that defines the continuous-time dynamics of the flow $D_\theta$ conditioned on $z$ and $\gG$. Note that, given the true distances $\vd(t_1)=\vd$, $\vd(t_0)$ can also be easily computed by reversing the continuous dynamics $D_\theta$: $D_\theta^{-1}(z, \gG)=\vd(t_1)+\int_{t_{1}}^{t_{0}} g_{\theta}(\vd(t), t, z, \gG) \diff t$. And thus the exact conditional log-likelihood of distances given $\gG$ can be computed by:
\begin{equation}
\label{eq:nll_distance}
\begin{aligned}
    \mathcal{L}_{aux} & =  \log{p_\theta(\vd|z,\gG)} \\
    &= \log{p(\vd(t_0))} - \int_{t_0}^{t_1} \operatorname{Tr}\left(\frac{\partial g_{\theta}}{\partial \vd (t)}\right)\diff t.
\end{aligned}
\end{equation}
An ODE solver can then be applied to estimate the gradients on parameters for optimization. In practice, $\gL_{aux}$ can be taken as an auxiliary objective defined on distances to supervise the training. In summary, the training objective can be interpreted as the sum of three parts:
\begin{equation}
\label{eq:loss}
    \gL(\theta,\phi,\psi) = \gL_{recon} + \lambda \gL_{prior} + \alpha \gL_{aux},
\end{equation}
where $\lambda$ and $\alpha$ are hyperparameters to reweight each component. The overall framework is illustrated in Fig.~\ref{fig:model}.

%% file: 32-training.tex
\subsection{End-to-end Learning via Hypergradient Descent}
\label{subsec:model-learning}


We now discuss how to optimize the bilevel problem defined by \eqref{eq:inner-loop} and \eqref{eq:outer-loop} through a practical algorithm. The inner problem in \eqref{eq:inner-loop} is a classic distance geometry problem about how to infer 3D coordinates from pairwise distances~\cite{,anand2018generative,simm2019generative,xu2021learning}. Others have used a semi-definite program (SDP) to infer protein structure from nuclear magnetic resonance data~\cite{alipanahi2013determining}, or an Alternating Direction Method of Multipliers (ADMM) algorithm to fold the protein into the 3D Cartesian coordinates~\cite{anand2018generative}. In this initial work we choose gradient descent (GD), with tractable learning dynamics $\Phi$, to approximately solve for the geometry:
\begin{equation}
    \mR_{{\theta,\phi}, t+1} = \Phi(\mR_{{\theta,\phi}, t}, \vd_{\theta,\phi}) = \mR_{{\theta,\phi}, t} - \eta \nabla H(\mR_{{\theta,\phi}, t},\vd_{\theta,\phi}),
\end{equation}
where $\eta$ is the learning rate and $\vd_{\theta,\phi}$ is the distance set generated from the distance prediction model. Under appropriate assumptions and for a number of updates $t \rightarrow \infty $, GD can converge to a proper geometry $\mR_{\theta,\phi}$ that depends on the predicted pairwise distances~\cite{bottou2010large}.

Now we consider how to calculate the hypergradient $\nabla_{\theta,\phi} \mathbb{E}_{z \sim q_{\phi}(z | \mR, \mathcal{G})}\left[F(\mR_{\theta,\phi})\right]$ from the outer loop reconstruction objective (\eqref{eq:outer-loop})  to train the model. 
Let $\mR_{{\theta,\phi}, T}$ denote the conformation generated by approximately solving for the distance geometry with $T$ steps gradient descent. Now we can write the hypergradient as:
\begin{align}
    \nabla_{\theta,\phi} \mathbb{E}&_{z \sim q_{\phi}(z | \mR, \mathcal{G})}\left[F(\mR_{{\theta,\phi},T})\right]\\
    & = \mathbb{E}_{z \sim q_{\phi}(z | \mR, \mathcal{G})} \partial_{\mR}  \left[F(\mR_{{\theta,\phi},T})\right] \nabla_{\theta,\phi} \mR_{{\theta,\phi},T},\nonumber
\end{align}
where the gradient $\nabla_{\theta,\phi} \mR_{\theta,\phi,T}$ can be computed by fully unrolling the dynamics of inner loop from $R_T$ to $R_0$.
Specifically, 
in  the forward computation, successive geometries $\mR_{0,\cdots,T}$ resulting from the optimization dynamics are cached. In the backward call, the cached geometries are used to compute gradients in a series of Vector-Jacobian Products (VJPs). 
During the reverse computation, the gradient starting from the $\partial_{\mR_T}F$ can be propagated to the intermediate geometries $R_t$ through $\nabla_{\mR_t}\mR_{t+1}$:
\begin{equation}
\label{eq:dR_dR}
\begin{aligned}
    \nabla_{\mR_t}\mR_{t+1} &= \nabla_{\mR_t} \big(\mR_t - \eta \nabla_{\mR_t}H(\vd_{\phi, \theta}, \mR_t)\big)\\
    & = 1 - \eta \nabla^2_{\mR_t} H(\vd_{\phi, \theta}, \mR_t)
\end{aligned}
\end{equation}
where $\nabla^2_{\mR_t}$ denotes the Hessian \textit{w.r.t.} ${\mR_t}$. 
With iteratively computed derivatives $\nabla_{\mR_t} \mR_T$, the adjoints on $\vd_{\phi, \theta}$ can be computed in forms of VJPs and further backpropagated to the parameters of encoder $q_\phi$ and decoder $p_\theta$.
Formally, $ \nabla_\vd \mR_T$ is computed by:
\begin{equation}
\begin{aligned}
\nabla_{\vd_{\theta,\psi}}\mR_T & =
\sum_{t=T-1}^0 [\nabla_{\mR_{t+1}} \mR_T] \nabla_{\vd} \mR_{t+1}\\
& = - \eta \sum_{t=T-1}^0 [\nabla_{\mR_{t+1}} \mR_T] \nabla_{\vd} \big (\nabla_{\mR_t}H(\vd_{\phi, \theta}, \mR_t) \big),
\end{aligned}
\end{equation}
where $\nabla_{\mR_{t+1}} \mR_T$ can be substituted by \eqref{eq:dR_dR}. The computation can be done efficiently with reverse-mode automatic differentiation software such as PyTorch~\cite{paszke2019pytorch}. A schematic illustration of the forward and backward computational graph through distance geometry is presented in Fig.~\ref{fig:inner}. We provide a detailed algorithm of the training procedure in Appendix.~\ref{app:sec:alg}.



\subsection{Sampling}
\label{subsec:model-sampling}

Given the graph $\gG$, to generate a conformation $\mR$, we first draw the latent variable $\tilde{z}$ from the prior distribution $p_\psi(z|\gG)$. Then we sample the random initial distances $\vd(t_0)$ from a Gaussian distribution, then pass $\tilde{\vd}(t_0)$ through the invertible Neural ODE $G_\theta$ conditioned on $\tilde{z}$ and $\gG$ to obtain the distance set $\tilde{\vd}=G_\theta(\tilde{\vd}(t_0);z,\gG)$. Then we produce the conformation $\mR$ by solving the distance geometry optimization problem $\arg\min_\mR H(\mR, \vd_{\theta,\phi})$ as defined in \eqref{eq:implicitdg}.

%% file: 40-exp.tex
\begin{table*}[!t]
\caption{Comparison of different methods on the conformation generation task. Top $5$ rows: deep generative models for molecular conformation generation. Bottom $6$ rows: different methods with an additional rule-based force field to further optimize the generated structures. We report the COV and MAT scores, where \textbf{Mean} and \textbf{Median} are calculated over different molecular graphs in the test set of GEOM. In practice, the size of the generated set is sampled as two times of the reference set following \citet{xu2021learning}.}
\label{table:confgen}
\begin{center}
\begin{threeparttable}[b]
    \input{tables/eval-confgen}
    \begin{tablenotes}
    	\item[*] For COV, the threshold $\delta$ is set as $0.5\mathrm{\AA}$ for QM9 and $1.25\mathrm{\AA}$ for Drugs following \citet{xu2021learning}. 
    \end{tablenotes}
\end{threeparttable}
\end{center}
\vspace{-10pt}
\end{table*}

\section{Experiments}

\subsection{Experiment Setup}

\textbf{Evaluation Tasks.} Following previous work on conformation generation~\cite{mansimov19molecular,simm2019generative,xu2021learning}, we conduct extensive experiments by comparing our method with the state-of-the-art baseline models on several standard tasks. \textbf{Conformation Generation} is formulated by \citet{xu2021learning}, who concentrate on the models' capacity to generate realistic and diverse molecular conformations. \textbf{Distance distribution modeling} is first proposed by \citet{simm2019generative}, who evaluate whether the methods can model the underlying distribution of distances.

\textbf{Baselines.} We compared our proposed model with the following state-of-the-art conformation generation methods. \textbf{CVGAE}~\citep{mansimov19molecular} is a conditional VAE-based model, which applied a few layers of graph neural networks to learn the atom representation from the molecular graph, and then directly predicts the 3D coordinates. \textbf{GraphDG}~\citep{simm2019generative} also employs the conditional VAE framework. Instead of directly generating the conformations in 3D coordinates, they instead learn the distribution over distances. Then the distances are converted into conformations with a distance geometry algorithm. \textbf{CGCF}~\cite{xu2021learning}, another two-stage method, uses continuous normalizing flows to predict the atomic pairwise distances. Following the baselines, we also compare our model with \textbf{RDKit}~\citep{riniker2015better}, a classical distance geometry approach built upon an extensive calculation collection of edge lengths by computational chemistry.

\textbf{Featurization and Implementation.} The MPNNs used for the encoder, prior and decoder are all implemented as Graph Isomorphism Networks~\cite{xu2018powerful,hu2019strategies}. For the input features of the graph representation, we only derive the atom and bond types from molecular graphs. As a default setup, the MPNNs are all implemented with $3$ layers, and the hidden embedding dimension is set as $256$. For the training of \modelname, we train the model on a single Tesla V100 GPU with a batch size of $128$ and a learning rate of $0.001$ until convergence, with Adam~\cite{kingma2013auto} as the optimizer. 

%% file: tables/eval-confgen.tex
\begin{tabular}{l | cc cc | cc cc}
\toprule
Dataset & \multicolumn{4}{c|}{GEOM-QM9} & \multicolumn{4}{c}{GEOM-Drugs} \\

\multirow{2}{*}{Metric} & \multicolumn{2}{c}{COV$^*$ (\%)} & \multicolumn{2}{c|}{MAT (\AA)} & \multicolumn{2}{c}{COV$^*$ (\%)} & \multicolumn{2}{c}{MAT (\AA)} \\

 & Mean & Median & Mean & Median & Mean & Median & Mean & Median \\ \midrule   



CVGAE & 8.52 & 5.62 & 0.7810 & 0.7811 &
        0.00 & 0.00 & 2.5225 & 2.4680 \\

GraphDG & 55.09 & 56.47 & 0.4649 & 0.4298 & 
          7.76 & 0.00 & 1.9840 & 2.0108 \\


CGCF & 69.60 & 70.64 & 0.3915 & 0.3986 & 49.92 & 41.07 & 1.2698 & 1.3064 \\


\modelname- & 75.57 & 80.76 & 0.3873 & 0.3850 & 51.24 & 46.36 & 1.2487 & 1.2609 \\

\modelname & \bf 77.98 & \bf 82.82 & \bf 0.3778 & \bf 0.3770 & \bf 52.59 & \bf 56.41 & \bf 1.2330 & \bf 1.2270 \\


\midrule
\midrule

RDKit & 79.94 & \bf 87.20 & 0.3238 & 0.3195 & 
        65.43 & 70.00 & 1.0962 & 1.0877 \\

\midrule

CVGAE + FF & 63.10 & 60.95 & 0.3939 & 0.4297 &
        83.08 & 95.21 & 0.9829 & 0.9177 \\

GraphDG + FF & 70.67 & 70.82 & 0.4168 & 0.3609 &
          84.68 & 93.94 & 0.9129 & 0.9090 \\


CGCF + FF & 73.52 & 72.75 & 0.3131 & 0.3251 & \bf 92.28 & 98.15 & 0.7740 & \bf 0.7338 \\


\modelname- + FF & 77.95 & 79.14 & 0.2851 & 0.2817 & 91.48 & 99.21 & 0.7743 & 0.7436\\
\modelname + FF & \bf 81.46 & \bf 83.80 & \bf 0.2702  & \bf 0.2709 & 91.88 & \bf 100.00 & \bf 0.7634 & \bf 0.7312 \\

\bottomrule
\end{tabular}

%% file: 41-conf-gen.tex
\subsection{Conformation Generation}
\label{subsec:conf-gen}

\begin{figure*}[!t]
\begin{center}
\resizebox{1.0\linewidth}{!}{
\includegraphics[width=\linewidth]{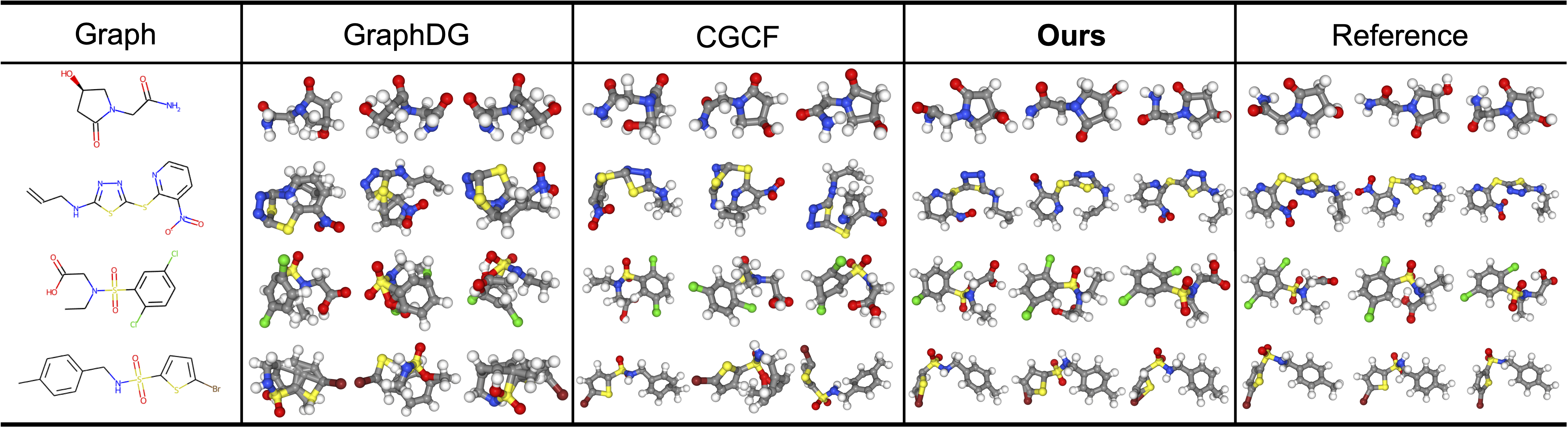}
}
\end{center}
\caption{Visualization of generated conformations from state-of-the-art baselines, our method and the reference set, where four molecular graphs are randomly taken from the test set of GEOM-Drugs. C, O, H, S and Cl are colored gray, red, white, yellow and green respectively.}
\label{fig:visualization}
\vspace{-5pt}
\end{figure*}

\textbf{Datasets.} Following \citet{xu2021learning}, we use the recent proposed GEOM-Drugs and GEOM-Drugs ~\citep{axelrod2020geom} datasets for the conformation generation task. 
The Geometric Ensemble Of Molecules (GEOM) dataset contains millions of high-quality stable conformations, which is suitable for the conformation generation task. 
The \textbf{GEOM-Drugs} dataset consists of generally medium-sized organic compounds, containing an average of 44.2 atoms.
We follow the setting from \citet{xu2021learning} to randomly take 50000 conformation-molecule pairs as the training set, and another 9161 conformations (covering 100 molecular graphs) as the test split. By contrast, \textbf{GEOM-QM9} is a much smaller dataset limited to small molecules with 9 heavy atoms.
Similarly, we randomly draw 50000 conformation-molecule pairs to constitute the training set, and another 17813 conformations covering 150 molecular graphs as the test set.

\textbf{Evaluation metrics.} In this task we hope the generated samples to be of both
high \textit{quality} and \textit{diversity}.
We follow previous work~\citep{hawkins2017conformation,mansimov19molecular,xu2021learning} to calculate the RMSD of the heavy atoms between generated samples and reference ones. Given the generated conformation $\mR$ and the reference $\mR^*$, we take the same alignment function $A(\mR,\mR^*)$ defined in \eqref{eq:rmsd} to obtain the aligned conformation $\hat{\mR}$, and then calculate the evaluation metric by $\operatorname{RMSD}(\mR, \hat{\mR})=\Big(\frac{1}{n} \sum_{i=1}^{n}\|\mR_{i}-\hat{\mR_{i}}\|^{2}\Big)^\frac{1}{2}$,
where $n$ is the number of heavy atoms. Built upon the RMSD metric, \citet{xu2021learning} defined \textbf{Cov}erage (COV) and \textbf{Mat}ching (MAT) scores to measure the diversity and quality respectively. 
COV counts the fraction of conformations in the reference set that are covered by at least one conformation in the generated set:
\begin{equation}
\label{eq:precision}
\begin{aligned}
    \operatorname{COV} (&\sS_g(\gG), \sS_r(\gG)) =\\
    &\frac{1}{| \sS_r |} \Big| \Big\{ \mR \in \sS_r  \big| \operatorname{ RMSD}(\mR, \mR') < \delta, \exists \mR' \in \sS_g  \Big\} \Big|.
\end{aligned}
\end{equation}
where $\sS_g(\gG)$ and $\sS_r(\gG)$ denote the generated and the reference conformations set respectively.
Typically, a higher COV score indicates a better diversity performance
to cover the complex true distribution.

While COV is able to detect mode-collapse, there is no guarantee for the quality of generated samples. 
Thus, the MAT score is defined as a complement metric that concentrates on the quality~\cite{xu2021learning}:
\begin{equation}
\label{eq:minmatchdist}
    \operatorname{MAT} (\sS_g(\gG), \sS_r(\gG)) = \frac{1}{| \sS_r|} \sum_{\mR' \in \sS_r } \min_{\mR \in \sS_g} \operatorname{RMSD}(\mR, \mR').
\end{equation}
Generally, more realistic generated samples lead to a lower MAT score.

\textbf{Results.} We calculate the COV and MAT evaluations on both GEOM-QM9 and GEOM-Drugs datasets for all baselines, and summarize the results in Tab.~\ref{table:confgen}. We visualize several representative examples in Fig.~\ref{fig:visualization}. Our \modelname outperforms all existing strong baselines with an obvious margin (top $5$ rows). By incorporating an end-to-end training objective via bilevel optimization, we consistently achieved a better result on all four metrics. By contrast, current state-of-the-art models GraphDG and CGCF 
suffer much worse performance due to the two-stage generation process, where the extra error caused by the distance geometry cannot be taken into account during training. CVGAE enjoys the same training and testing objective, but still shows inferior performance since it fails to keep the vital translation and rotation invariant property.

Similar to previous work~\cite{mansimov19molecular,xu2021learning}, we also further test all models by incorporating a rule-based empirical force field~\citep{halgren1996merck-extension} and compare the performance with the classic RDKit toolkit. Specifically, we first generate the conformations with the generative models as initial structures, and then utilize the force field to further optimize the generated structures. The additional results are reported in Tab.~\ref{table:confgen} (bottom $6$ rows). As shown in the table, \modelname still achieves the best results among all generative models. More importantly, our method outperforms RDKit on $7$ out of $8$ evaluations and achieves competitive results on the other one, making our method the first generative model that already practically useful for real-world applications.

\textbf{Ablation Study.} So far we have demonstrated the superior performance of the proposed method. However, because we adopt a slightly different architecture, it remains unclear where the effectiveness comes from. In this part, we carefully conduct an ablation study by removing the bilevel component defined in \eqref{eq:outer-loop} during training, \ie, remove $\gL_{recon}$ and learn the model with only $\gL_{aux}$ and $\gL_{prior}$. We denote this variant of \modelname as \modelname-. and summarize the additional results in Tab.~\ref{table:confgen}.

As shown in the table, removing the bilevel component hurts performance. These results verify that only learning from distances will introduce an extra bias for the generated conformations, and our end-to-end method for directly learning on the 3D structure helps to overcome this issue. Another observation is that as a combination of flow-based and VAE-based model, \modelname- still achieves significantly better results than the Flow-based CGCF and VAE-based GraphDG, with exactly the same training and sampling process. This result indicates that incorporating both global ($z$) and local $\vd(t_0)$ latent variables will contribute to the generated conformations, which can help to capture both the global and local geometric structure and atomic interactions. 

%% file: 42-distance.tex
\begin{table}[!t]
\caption{Comparison of different models on the distance distribution modeling task. We compare the marginals ($p(d_{uv}|\gG)$), pairs ($p(d_{uv},d_{ij}|\gG)$) and joint distribution ($p(\vd|\gG)$) of edges connecting C and O atoms. We report the \textbf{Median} and \textbf{Mean} of the MMD metric. Molecular graphs $\gG$ are taken from the test set of ISO17.}
\label{table:distance}
\begin{center}
\vspace{-5pt}
\resizebox{\linewidth}{!}{
\input{tables/eval-distance}
}
\end{center}
\vspace{-8pt}
\end{table}

\subsection{Distance Distribution Modeling}
\label{subsec:dist}

\textbf{Dataset.} For the distances modeling task, we follow \citet{simm2019generative,xu2021learning} and use the ISO17 dataset~\citep{simm2019generative}. ISO17 is constructed from the snapshots of \textit{ab initio} molecular dynamics simulations, where the coordinates are not just equilibrium conformations but are samples that reflect the underlying density around equilibrium states. 
We follow previous work to split ISO17 into a training set with 167 molecules and a test set with the other 30 molecules.

\textbf{Evaluation metrics.} To obtain a distribution over distances from a distribution over conformations, we sample a set of conformations $R$ and then calculate the corresponding atomic lengths between C and O atoms (H atoms are usually ignored).
Let $p(d_{uv}|\gG)$ denote the conditional distribution of distances on each edge $e_{uv}$ given a molecular graph $\gG$.
To evaluate the distance distributions, 
we use the maximum mean discrepancy (MMD)~\citep{gretton2012kernel} to the ground-truth distributions. More specifically, we evaluate against
the ground truth the MMD of marginal distributions of each individual edge's distance $p(d_{uv} | \gG)$, pairs of distances $p(d_{uv}, d_{ij} | \gG)$ and the joint distance $p(\vd | \gG)$. For this benchmark, the size of the generated sample set is the same as the reference set.

\textbf{Results.} The results of MMDs are summarized in Tab.~\ref{table:distance}. The statistics show that the generated distance distribution of \modelname is significantly closer to the ground-truth distribution compared with the baseline models. These results demonstrate that our method can not only generate realistic conformations, but also model the density around equilibrium states. By contrast, though RDKit shows competitive performance for conformation generation, it seems to struggle with the distribution modeling benchmark. This is because RDKit is only designed to find the equilibrium states by using the empirical force field~\cite{halgren1996merck}, and thus it lacks the capacity to capture the underlying distribution. The further ablation study between \modelname and \modelname- also verifies the effectiveness of the bilevel optimization components.






%% file: tables/eval-distance.tex
\begin{tabular}{l cc cc cc}
\toprule
 & \multicolumn{2}{c}{Single} & \multicolumn{2}{c}{Pair} & \multicolumn{2}{c}{All} \\

 & Mean & Median & Mean & Median & Mean & Median \\
\midrule

RDKit & 
3.4513 & 3.1602 & 
3.8452 & 3.6287 & 
4.0866 & 3.7519 \\

CVGAE & 
4.1789 & 4.1762 & 
4.9184 & 5.1856 & 
5.9747 & 5.9928 \\

GraphDG & 
0.7645 & 0.2346 & 
0.8920 & 0.3287 & 
1.1949 & 0.5485 \\

CGCF    & 
0.4490 & 0.1786 & 
0.5509 & 0.2734 & 
0.8703 & 0.4447 \\ \midrule

\modelname-    & 
0.2551 & 0.1352 & 
0.2719 & 0.1742 & 
0.2968 & 0.2132 \\

\modelname    & 
\bf 0.1809 & \bf 0.1153 & 
\bf 0.1946 & \bf 0.1455 & 
\bf 0.2113 & \bf 0.2047 \\


\bottomrule
\end{tabular}

%% file: 50-related.tex
\section{Related Work}

In recent years, deep learning has shown significant progress for 3D structure generation.
There have been works using neural networks to derive energy prediction models, which then are taken as faster alternatives to quantum mechanics-based energy calculations~\citep{schutt2017schnet,smith2017ani}
for molecular dynamics simulation or molecule optimization~\cite{Wang2020SIL}. However,
though accelerated by neural networks, these approaches are still time-consuming due to the lengthy sampling process.
Recently, \cite{gebauer2019symmetry} and \cite{hoffmann2019generating} provide methods to generate new 3D molecules with deep generative models, while \cite{simm2020reinforcement} and \cite{simm2020symmetry} employ reinforcement learning to search the vast geometric space. However, none of these methods is designed to generate the conformations from the molecular graph structure, making them orthogonal to our framework. \cite{gogineni2020torsionnet} proposes TorsionNet, which uses RL for conformation search by determining torsional angles, and takes a classical force field for state transition and reward evaluation. However, this model is specifically designed for larger molecules, and incapable of modeling other complex geometric structures such as bond angles and lengths. Therefore, it is also not comparable in our setting.

Many other works~\citep{lemke2019encodermap,alquraishi2019protein, ingraham2019protein,noe2019boltzmann} also learn to directly predict 3D structures, but focus on the protein folding problem. Specifically, \citet{senior2020protein,jumper2020high} significantly advance this field with an end-to-end attention-based model called AlphaFold. Unfortunately, proteins are amino-acid sequences with low chemical diversity, much larger scale and for which abundant structural exists while general molecules are highly structured graphs with a variety of cycles and much broader chemical composition, making it unclear whether these methods are transferable to the general conformation generation task. 




%% file: 60-conclusion.tex
\section{Conclusion}

In this paper, we propose \modelname, an end-to-end framework for molecular conformation generation via bilevel programming. Our generative model can overcome significant errors of previous two-stage models, thanks to the end-to-end training based on bilevel programming, while keeping the property of rotational and translational invariance.
Experimental results demonstrate the superior performance of our method over all state-of-the-art baselines on several standard benchmarks. 
Future work includes combining our bilevel optimization framework with other kinds of generative models, and extending our method to other challenging structures such as proteins.

%% file: 65-acknowledgment.tex
\section*{Acknowledgments}

This project is supported by the Natural Sciences and Engineering Research Council (NSERC) Discovery Grant, the Canada CIFAR AI Chair Program, collaboration grants between Microsoft Research and Mila, Samsung Electronics Co., Ldt., Amazon Faculty Research Award, Tencent AI Lab Rhino-Bird Gift Fund and a NRC Collaborative R\&D Project (AI4D-CORE-06). This project was also partially funded by IVADO Fundamental Research Project grant PRF-2019-3583139727.

%% file: 70-app.tex
\appendix

\section{Data Preprocess}
\label{app:sec:preprocess}

Inspired by classic molecular distance geometry~\citep{crippen1988distance}, in our framework we also generate the confirmations by taking the inter-atomic distances as the intermediate variables,
which enables the invariant property to rotation and translation. In practice, the chemical bonds existing in the molecular graph are not sufficient to determine a conformation, and thus we follow existing works~\cite{simm2019generative,xu2021learning} to first expand the graphs by extending \textit{auxiliary} edges. Specifically, the atoms that are $2$ or $3$ hops away are connected with \textit{virtual bonds}, labeled differently from the real bonds in the vanilla molecule. These extra bonds contribute to reducing the degrees of freedom in the 3D coordinates and characterizing the unique graph, with the edges between $2$-hop neighbors helping to fix the angles between atoms, and those between $3$-hop neighbors fixing dihedral angles.

\section{Training Algorithm}
\label{app:sec:alg}

\begin{algorithm}[!h]
  \caption{Training Algorithm of \modelname.}
  \label{alg:pointflow-train}
\textbf{Input}: objective reweighting coefficients $\alpha$ and $\lambda$; the inner loop optimization iterations $T$ and learning rate $\eta$; alignment function $A(\cdot,\cdot)$; data samples $\{\gG_t, \mR^*_t\}$.\\
\textbf{Initial}: prior $p_\psi(z|\gG)$, decoder $p_\theta(\mR|z,\gG)$ and its dynamics defined as $g_\theta$, encoder $q_\phi(z|\mR,\gG)$
\begin{algorithmic}
\WHILE{$\theta,\phi,\psi$ have not converged}
    \STATE {$\mu,\sigma \leftarrow q_\phi(z|\gG_t,\mR^*_t)$} 
    \STATE {$z \leftarrow \epsilon \odot \sigma + \mu $} \hfill \COMMENT{Reparameterization}
    \STATE {$\mu_q,\sigma_q \leftarrow p_\psi(z|\gG_t)$} 
    \STATE {$\mathcal{L}_{prior} = \frac{1}{2}\log \frac{\sigma}{\sigma_q} - \frac{\sigma_q^2+(\mu_q-\mu)^2}{2\sigma^2} $}
    \STATE $\vd^* \leftarrow \mR^*_t$ \hfill \COMMENT{Calculate $\vd$ from $\mR^*$}
    \STATE $\vd^*_0 = D_{\theta}^{-1}(z, \gG) = \vd^* + \int_{t_1}^{t_0} g_\theta(\vd^*(t), t , \gG, z) \diff t $
    \STATE $\mathcal{L}_{aux} = \log{p(\vd^*_0)} - \int_{t_0}^{t_1} \operatorname{Tr}\left(\frac{\partial g_{\theta}}{\partial \vd (t)}\right)\diff t$
    \STATE Initialize $\mR_0$, sample $\vd(t_0) \sim \gN(\vzero, \mI)$
    \STATE $\vd = D_\theta(z,\gG) = \vd(t_0) + \int_{t_0}^{t_1} g_\theta(\vd(t), t , \gG, z) \diff t$
    \FOR{$t = 1, 2, \cdots, T$}
        \STATE $\mR_{t+1} = \mR_{t} - \eta \nabla H(\mR_{t},\vd)$ \hfill \COMMENT{Inner loop}
    \ENDFOR 
    \STATE $\mR \leftarrow \mR_T$
    \STATE {$\mathcal{L}_{recon} = -\sum_{i=1}^{n} \sum_{j=1}^{3}\left(\mR_{ij}-A(\mR, \mR^*)_{ij}\right)^{2}$}
    \STATE {$\mathcal{L} = \gL_{recon} + \lambda \gL_{prior} + \alpha \gL_{aux}$}
    \STATE {$\theta, \phi, \psi \leftarrow \operatorname{Adam}(\mathcal{L}; \theta, \phi, \psi)$}
\ENDWHILE
\STATE 
\textbf{return} $q_\phi$, $p_\theta$, $p_\psi$
\end{algorithmic}
\end{algorithm}

\section{Additional Comparisons}

\subsection{Property Prediction}

This task is first proposed in \citet{simm2019generative}, which estimates the expected molecular properties for molecular graphs by a set of generated conformations. This task can further demonstrate the effectiveness and quality of generated samples, and is important for many real-world applications such as drug and material design. 

\textbf{Dataset.} Following \citet{simm2019generative}, we also employ the ISO17 dataset. More details about the dataset can be found in Sec.~\ref{subsec:dist}.

\textbf{Evaluation metrics.} For comparison, we calculate the ensemble properties of each molecular graph by averaging over a set of generated conformations. Specifically, we calculate the total electronic energy $E_\text{elec}$, the energy of HOMO $\epsilon_\text{HOMO}$ and the LUMO $\epsilon_\text{LUMO}$, and the dipole moment $\mu$, using the quantum chemical calculation package Psi4~\cite{smith2020psi4}. In practice, we generate $50$ samples from different methods to estimate the property, and report median error of averaged properties to measure the accuracy of predicted properties. Similar to \citet{simm2019generative}, we exclude CVGAE from this analysis due to its poor generated quality. 

\textbf{Results.} The results are shown in Tab.~\ref{app:table:property}. As shown in the table, \modelname outperforms all other generative models, and shows competitive results compared with RDKit. Close observation indicates that CGCF struggles with this task since the generated conformations suffer a extremely high variance. By contrast, our proposed method enjoys the best performance thanks to the high quality of generated samples.

\begin{table}[!ht]
\caption{Median difference in averaged properties between ground-truth and generated conformations from different methods. Unit: $E_\text{elec}$(kJ/mol), $\epsilon_\text{HOMO}$(eV), $\epsilon_\text{LUMO}$(eV), $\mu$(debye).}
\label{app:table:property}
\begin{center}
\input{tables/eval-prop}
\end{center}
\end{table}


\subsection{More Results of Coverage Score}

\begin{figure*}[!t]
    \centering
    \includegraphics[width=0.99\textwidth]{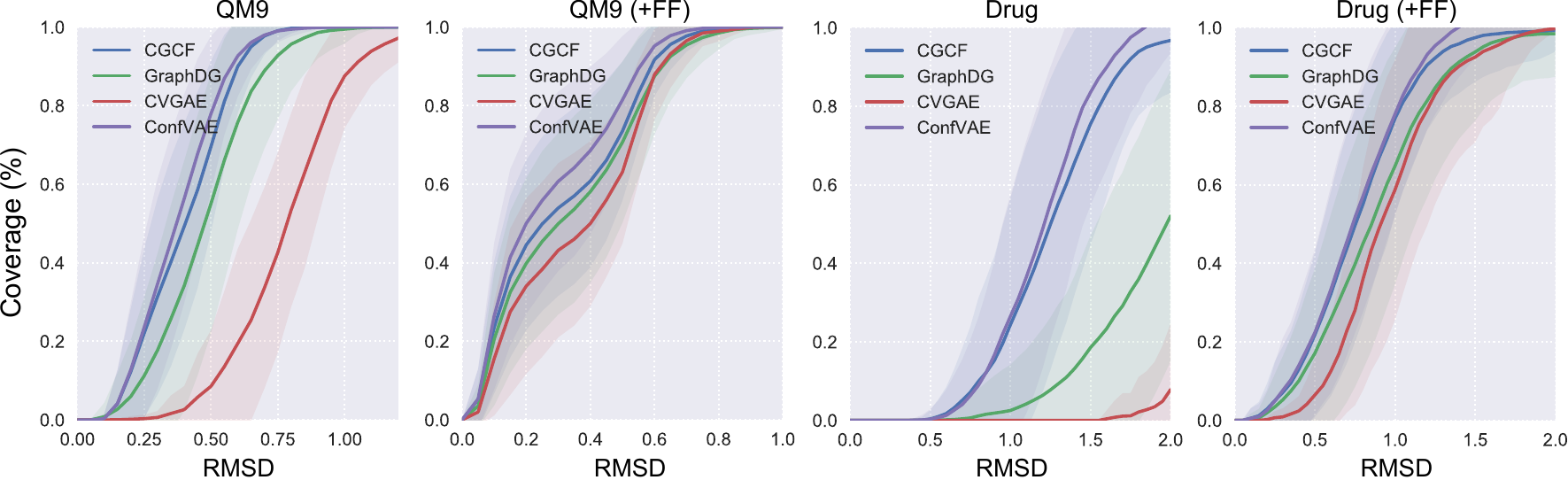}
    \caption{Curves of the \textbf{Cov}erage score with different thresholds $\delta$ on GEOM-QM9 (left two) and GEOM-Drugs (right two) datasets. The first and third curves evaluates the generated conformations from different generative models, while the other two are further optimized with the empirical force field.}
    \label{app:fig:coverage}
\end{figure*}

\begin{figure*}[!ht]
\begin{center}
    \centering
    \includegraphics[width=0.99\textwidth]{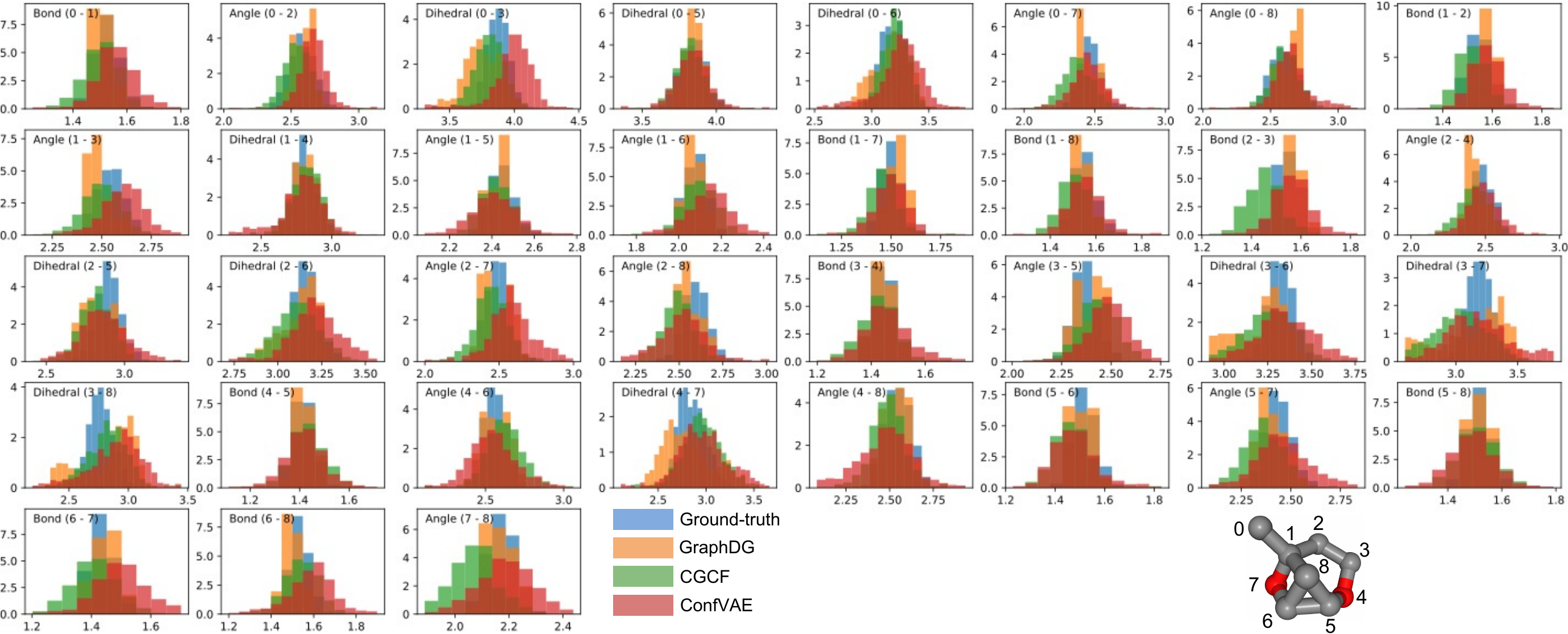}
\end{center}
\caption{Marginal distributions $p(d_{uv}|\gG)$ of ground-truth and generated conformations from generative models. We study the edges between C and O atoms, and omit the H atoms for clarity. In each subplot, the annotation ($u-v$) denotes the corresponding atoms connected by the chemical bond $d_{uv}$.}
\label{app:fig:distdistrib}
\end{figure*}

In this section, we give more results about \textbf{Cov}erage score with different thresholds $\delta$. The details about the COV score can be found in Sec.~\ref{subsec:conf-gen}. Results are shown in Fig.~\ref{app:fig:coverage}. As shown in the figure, \modelname consistently achieves better performance than previous state-of-the-art models, which demonstrates our proposed method is capable to generate more realistic samples.

\subsection{Visualization of Distributions}

In Fig.~\ref{app:fig:distdistrib}, we investigate the accuracy of generated conformations by visualizing the marginal distributions $p(d_{uv}|\gG)$ for all pairwise distances between C and O atoms of a molecular graph in the ISO17 test set. As shown in the figure, though primarily designed for learning the 3D structures via an end-to-end framework, our method can still make a much better estimation of the distance distributions than the state-of-the-art model for molecular geometry modeling. As a representative element of the pairwise property between atoms, the inter-atomic distances demonstrate the capacity of our model to capture the inter-atomic interactions.


%% file: tables/eval-prop.tex
\begin{tabular}{l cccc}
\toprule
& $E_{elec}$ & $\epsilon_\text{HOMO}$ & $\epsilon_\text{LUMO}$ & $\mu$ \\
\midrule

RDKit & 42.7 & \bf 0.08 & 0.15 & \bf 0.29 \\

GraphDG & 58.0 & 0.10 & 0.09 & 0.33 \\

CGCF & 208.2 & 0.80 & 1.11 & 0.46 \\

\bf \modelname & \bf 40.2 & 0.10 & \bf 0.08 & \bf 0.29 \\

\bottomrule
\end{tabular}

%% file: icml2021.bbl
\begin{thebibliography}{56}
\providecommand{\natexlab}[1]{#1}
\providecommand{\url}[1]{\texttt{#1}}
\expandafter\ifx\csname urlstyle\endcsname\relax
  \providecommand{\doi}[1]{doi: #1}\else
  \providecommand{\doi}{doi: \begingroup \urlstyle{rm}\Url}\fi

\bibitem[Alipanahi et~al.(2013)Alipanahi, Krislock, Ghodsi, Wolkowicz,
  Donaldson, and Li]{alipanahi2013determining}
Alipanahi, B., Krislock, N., Ghodsi, A., Wolkowicz, H., Donaldson, L., and Li,
  M.
\newblock Determining protein structures from noesy distance constraints by
  semidefinite programming.
\newblock \emph{Journal of Computational Biology}, 20\penalty0 (4):\penalty0
  296--310, 2013.

\bibitem[AlQuraishi(2019)]{alquraishi2019protein}
AlQuraishi, M.
\newblock End-to-end differentiable learning of protein structure.
\newblock \emph{Cell systems}, 8\penalty0 (4):\penalty0 292--301, 2019.

\bibitem[Anand \& Huang(2018)Anand and Huang]{anand2018generative}
Anand, N. and Huang, P.-S.
\newblock Generative modeling for protein structures.
\newblock In \emph{Proceedings of the 32nd International Conference on Neural
  Information Processing Systems}, pp.\  7505--7516, 2018.

\bibitem[Axelrod \& Gomez-Bombarelli(2020)Axelrod and
  Gomez-Bombarelli]{axelrod2020geom}
Axelrod, S. and Gomez-Bombarelli, R.
\newblock Geom: Energy-annotated molecular conformations for property
  prediction and molecular generation.
\newblock \emph{arXiv preprint arXiv:2006.05531}, 2020.

\bibitem[Ballard et~al.(2015)Ballard, Martiniani, Stevenson, Somani, and
  Wales]{ballard2015exploiting}
Ballard, A.~J., Martiniani, S., Stevenson, J.~D., Somani, S., and Wales, D.~J.
\newblock Exploiting the potential energy landscape to sample free energy.
\newblock \emph{Wiley Interdisciplinary Reviews: Computational Molecular
  Science}, 5\penalty0 (3):\penalty0 273--289, 2015.

\bibitem[Bengio(2000)]{bengio2000gradient}
Bengio, Y.
\newblock Gradient-based optimization of hyperparameters.
\newblock \emph{Neural computation}, 12\penalty0 (8):\penalty0 1889--1900,
  2000.

\bibitem[Bennett et~al.(2006)Bennett, Hu, Ji, Kunapuli, and
  Pang]{bennett2006model}
Bennett, K.~P., Hu, J., Ji, X., Kunapuli, G., and Pang, J.-S.
\newblock Model selection via bilevel optimization.
\newblock In \emph{The 2006 IEEE International Joint Conference on Neural
  Network Proceedings}, pp.\  1922--1929. IEEE, 2006.

\bibitem[Bottou(2010)]{bottou2010large}
Bottou, L.
\newblock Large-scale machine learning with stochastic gradient descent.
\newblock In \emph{Proceedings of COMPSTAT'2010}, pp.\  177--186. Springer,
  2010.

\bibitem[Bruna et~al.(2013)Bruna, Zaremba, Szlam, and LeCun]{bruna2013spectral}
Bruna, J., Zaremba, W., Szlam, A., and LeCun, Y.
\newblock Spectral networks and locally connected networks on graphs.
\newblock \emph{arXiv preprint arXiv:1312.6203}, 2013.

\bibitem[Chen et~al.(2018)Chen, Rubanova, Bettencourt, and
  Duvenaud]{chen2018neural}
Chen, R.~T., Rubanova, Y., Bettencourt, J., and Duvenaud, D.~K.
\newblock Neural ordinary differential equations.
\newblock In \emph{Advances in neural information processing systems}, pp.\
  6571--6583, 2018.

\bibitem[Colson et~al.(2007)Colson, Marcotte, and Savard]{colson2007overview}
Colson, B., Marcotte, P., and Savard, G.
\newblock An overview of bilevel optimization.
\newblock \emph{Annals of operations research}, 153\penalty0 (1):\penalty0
  235--256, 2007.

\bibitem[Crippen et~al.(1988)Crippen, Havel, et~al.]{crippen1988distance}
Crippen, G.~M., Havel, T.~F., et~al.
\newblock \emph{Distance geometry and molecular conformation}, volume~74.
\newblock Research Studies Press Taunton, 1988.

\bibitem[De~Vivo et~al.(2016)De~Vivo, Masetti, Bottegoni, and
  Cavalli]{de2016role}
De~Vivo, M., Masetti, M., Bottegoni, G., and Cavalli, A.
\newblock Role of molecular dynamics and related methods in drug discovery.
\newblock \emph{Journal of medicinal chemistry}, 59\penalty0 (9):\penalty0
  4035--4061, 2016.

\bibitem[Domke(2012)]{domke2012generic}
Domke, J.
\newblock Generic methods for optimization-based modeling.
\newblock In \emph{Artificial Intelligence and Statistics}, pp.\  318--326,
  2012.

\bibitem[Duvenaud et~al.(2015)Duvenaud, Maclaurin, Aguilera-Iparraguirre,
  G{\'o}mez-Bombarelli, Hirzel, Aspuru-Guzik, and
  Adams]{duvenaud2015convolutional}
Duvenaud, D., Maclaurin, D., Aguilera-Iparraguirre, J., G{\'o}mez-Bombarelli,
  R., Hirzel, T., Aspuru-Guzik, A., and Adams, R.~P.
\newblock Convolutional networks on graphs for learning molecular fingerprints.
\newblock \emph{arXiv preprint arXiv:1509.09292}, 2015.

\bibitem[Flamary et~al.(2014)Flamary, Rakotomamonjy, and
  Gasso]{flamary2014learning}
Flamary, R., Rakotomamonjy, A., and Gasso, G.
\newblock Learning constrained task similarities in graphregularized multi-task
  learning.
\newblock \emph{Regularization, Optimization, Kernels, and Support Vector
  Machines}, pp.\  103, 2014.

\bibitem[Franceschi et~al.(2017)Franceschi, Donini, Frasconi, and
  Pontil]{franceschi2017forward}
Franceschi, L., Donini, M., Frasconi, P., and Pontil, M.
\newblock Forward and reverse gradient-based hyperparameter optimization.
\newblock \emph{arXiv preprint arXiv:1703.01785}, 2017.

\bibitem[Franceschi et~al.(2018)Franceschi, Frasconi, Salzo, Grazzi, and
  Pontil]{franceschi2018bilevel}
Franceschi, L., Frasconi, P., Salzo, S., Grazzi, R., and Pontil, M.
\newblock Bilevel programming for hyperparameter optimization and
  meta-learning.
\newblock \emph{arXiv preprint arXiv:1806.04910}, 2018.

\bibitem[Gebauer et~al.(2019)Gebauer, Gastegger, and
  Sch{\"u}tt]{gebauer2019symmetry}
Gebauer, N., Gastegger, M., and Sch{\"u}tt, K.
\newblock Symmetry-adapted generation of 3d point sets for the targeted
  discovery of molecules.
\newblock In \emph{Advances in Neural Information Processing Systems}, pp.\
  7566--7578, 2019.

\bibitem[Gilmer et~al.(2017)Gilmer, Schoenholz, Riley, Vinyals, and
  Dahl]{gilmer2017neural}
Gilmer, J., Schoenholz, S.~S., Riley, P.~F., Vinyals, O., and Dahl, G.~E.
\newblock Neural message passing for quantum chemistry.
\newblock \emph{arXiv preprint arXiv:1704.01212}, 2017.

\bibitem[Gogineni et~al.(2020)Gogineni, Xu, Punzalan, Jiang, Kammeraad, Tewari,
  and Zimmerman]{gogineni2020torsionnet}
Gogineni, T., Xu, Z., Punzalan, E., Jiang, R., Kammeraad, J., Tewari, A., and
  Zimmerman, P.
\newblock Torsionnet: A reinforcement learning approach to sequential conformer
  search.
\newblock \emph{arXiv preprint arXiv:2006.07078}, 2020.

\bibitem[Gretton et~al.(2012)Gretton, Borgwardt, Rasch, Sch{\"o}lkopf, and
  Smola]{gretton2012kernel}
Gretton, A., Borgwardt, K.~M., Rasch, M.~J., Sch{\"o}lkopf, B., and Smola, A.
\newblock A kernel two-sample test.
\newblock \emph{The Journal of Machine Learning Research}, 13\penalty0
  (1):\penalty0 723--773, 2012.

\bibitem[Griewank \& Walther(2008)Griewank and Walther]{griewank2008evaluating}
Griewank, A. and Walther, A.
\newblock \emph{Evaluating derivatives: principles and techniques of
  algorithmic differentiation}.
\newblock SIAM, 2008.

\bibitem[Halgren(1996{\natexlab{a}})]{halgren1996merck}
Halgren, T.~A.
\newblock Merck molecular force field. i. basis, form, scope, parameterization,
  and performance of mmff94.
\newblock \emph{Journal of computational chemistry}, 17\penalty0
  (5-6):\penalty0 490--519, 1996{\natexlab{a}}.

\bibitem[Halgren(1996{\natexlab{b}})]{halgren1996merck-extension}
Halgren, T.~A.
\newblock Merck molecular force field. v. extension of mmff94 using
  experimental data, additional computational data, and empirical rules.
\newblock \emph{Journal of Computational Chemistry}, 17\penalty0
  (5-6):\penalty0 616--641, 1996{\natexlab{b}}.

\bibitem[Hawkins(2017)]{hawkins2017conformation}
Hawkins, P.~C.
\newblock Conformation generation: the state of the art.
\newblock \emph{Journal of Chemical Information and Modeling}, 57\penalty0
  (8):\penalty0 1747--1756, 2017.

\bibitem[Hoffmann \& No{\'e}(2019)Hoffmann and No{\'e}]{hoffmann2019generating}
Hoffmann, M. and No{\'e}, F.
\newblock Generating valid euclidean distance matrices.
\newblock \emph{arXiv preprint arXiv:1910.03131}, 2019.

\bibitem[Hu et~al.(2019)Hu, Liu, Gomes, Zitnik, Liang, Pande, and
  Leskovec]{hu2019strategies}
Hu, W., Liu, B., Gomes, J., Zitnik, M., Liang, P., Pande, V., and Leskovec, J.
\newblock Strategies for pre-training graph neural networks.
\newblock \emph{arXiv preprint arXiv:1905.12265}, 2019.

\bibitem[Ingraham et~al.(2019)Ingraham, Riesselman, Sander, and
  Marks]{ingraham2019protein}
Ingraham, J., Riesselman, A.~J., Sander, C., and Marks, D.~S.
\newblock Learning protein structure with a differentiable simulator.
\newblock In \emph{International Conference on Learning Representations}, 2019.

\bibitem[Jumper et~al.(2020)Jumper, Evans, Pritzel, Green, Figurnov,
  Tunyasuvunakool, Ronneberger, Bates, Zidek, Bridgland,
  et~al.]{jumper2020high}
Jumper, J., Evans, R., Pritzel, A., Green, T., Figurnov, M., Tunyasuvunakool,
  K., Ronneberger, O., Bates, R., Zidek, A., Bridgland, A., et~al.
\newblock High accuracy protein structure prediction using deep learning.
\newblock \emph{Fourteenth Critical Assessment of Techniques for Protein
  Structure Prediction (Abstract Book)}, 22:\penalty0 24, 2020.

\bibitem[Kearnes et~al.(2016)Kearnes, McCloskey, Berndl, Pande, and
  Riley]{kearnes2016molecular}
Kearnes, S., McCloskey, K., Berndl, M., Pande, V., and Riley, P.
\newblock Molecular graph convolutions: moving beyond fingerprints.
\newblock \emph{Journal of computer-aided molecular design}, 30\penalty0
  (8):\penalty0 595--608, 2016.

\bibitem[Kingma \& Welling(2013)Kingma and Welling]{kingma2013auto}
Kingma, D.~P. and Welling, M.
\newblock Auto-encoding variational bayes.
\newblock \emph{arXiv preprint arXiv:1312.6114}, 2013.

\bibitem[Kipf \& Welling(2016)Kipf and Welling]{kipf2016semi}
Kipf, T.~N. and Welling, M.
\newblock Semi-supervised classification with graph convolutional networks.
\newblock \emph{arXiv preprint arXiv:1609.02907}, 2016.

\bibitem[Lemke \& Peter(2019)Lemke and Peter]{lemke2019encodermap}
Lemke, T. and Peter, C.
\newblock Encodermap: Dimensionality reduction and generation of molecule
  conformations.
\newblock \emph{Journal of chemical theory and computation}, 15\penalty0
  (2):\penalty0 1209--1215, 2019.

\bibitem[Liberti et~al.(2014)Liberti, Lavor, Maculan, and
  Mucherino]{liberti2014euclidean}
Liberti, L., Lavor, C., Maculan, N., and Mucherino, A.
\newblock Euclidean distance geometry and applications.
\newblock \emph{SIAM review}, 56\penalty0 (1):\penalty0 3--69, 2014.

\bibitem[Maclaurin et~al.(2015)Maclaurin, Duvenaud, and
  Adams]{maclaurin2015gradient}
Maclaurin, D., Duvenaud, D., and Adams, R.
\newblock Gradient-based hyperparameter optimization through reversible
  learning.
\newblock In \emph{International Conference on Machine Learning}, pp.\
  2113--2122, 2015.

\bibitem[Mansimov et~al.(2019)Mansimov, Mahmood, Kang, and
  Cho]{mansimov19molecular}
Mansimov, E., Mahmood, O., Kang, S., and Cho, K.
\newblock Molecular geometry prediction using a deep generative graph neural
  network.
\newblock \emph{arXiv preprint arXiv:1904.00314}, 2019.

\bibitem[Mu{\~n}oz-Gonz{\'a}lez et~al.(2017)Mu{\~n}oz-Gonz{\'a}lez, Biggio,
  Demontis, Paudice, Wongrassamee, Lupu, and Roli]{munoz2017towards}
Mu{\~n}oz-Gonz{\'a}lez, L., Biggio, B., Demontis, A., Paudice, A.,
  Wongrassamee, V., Lupu, E.~C., and Roli, F.
\newblock Towards poisoning of deep learning algorithms with back-gradient
  optimization.
\newblock In \emph{Proceedings of the 10th ACM Workshop on Artificial
  Intelligence and Security}, pp.\  27--38, 2017.

\bibitem[No{\'e} et~al.(2019)No{\'e}, Olsson, K{\"o}hler, and
  Wu]{noe2019boltzmann}
No{\'e}, F., Olsson, S., K{\"o}hler, J., and Wu, H.
\newblock Boltzmann generators: Sampling equilibrium states of many-body
  systems with deep learning.
\newblock \emph{Science}, 365\penalty0 (6457):\penalty0 eaaw1147, 2019.

\bibitem[Paszke et~al.(2019)Paszke, Gross, Massa, Lerer, Bradbury, Chanan,
  Killeen, Lin, Gimelshein, Antiga, et~al.]{paszke2019pytorch}
Paszke, A., Gross, S., Massa, F., Lerer, A., Bradbury, J., Chanan, G., Killeen,
  T., Lin, Z., Gimelshein, N., Antiga, L., et~al.
\newblock Pytorch: An imperative style, high-performance deep learning library.
\newblock \emph{arXiv preprint arXiv:1912.01703}, 2019.

\bibitem[Riniker \& Landrum(2015)Riniker and Landrum]{riniker2015better}
Riniker, S. and Landrum, G.~A.
\newblock Better informed distance geometry: using what we know to improve
  conformation generation.
\newblock \emph{Journal of chemical information and modeling}, 55\penalty0
  (12):\penalty0 2562--2574, 2015.

\bibitem[Scarselli et~al.(2008)Scarselli, Gori, Tsoi, Hagenbuchner, and
  Monfardini]{scarselli2008graph}
Scarselli, F., Gori, M., Tsoi, A.~C., Hagenbuchner, M., and Monfardini, G.
\newblock The graph neural network model.
\newblock \emph{IEEE transactions on neural networks}, 20\penalty0
  (1):\penalty0 61--80, 2008.

\bibitem[Sch{\"u}tt et~al.(2017)Sch{\"u}tt, Kindermans, Felix, Chmiela,
  Tkatchenko, and M{\"u}ller]{schutt2017schnet}
Sch{\"u}tt, K., Kindermans, P.-J., Felix, H. E.~S., Chmiela, S., Tkatchenko,
  A., and M{\"u}ller, K.-R.
\newblock Schnet: A continuous-filter convolutional neural network for modeling
  quantum interactions.
\newblock In \emph{Advances in neural information processing systems}, pp.\
  991--1001, 2017.

\bibitem[Senior et~al.(2020{\natexlab{a}})Senior, Evans, Jumper, Kirkpatrick,
  Sifre, Green, Qin, {\v{Z}}{\'\i}dek, Nelson, Bridgland,
  et~al.]{senior2020improved}
Senior, A.~W., Evans, R., Jumper, J., Kirkpatrick, J., Sifre, L., Green, T.,
  Qin, C., {\v{Z}}{\'\i}dek, A., Nelson, A.~W., Bridgland, A., et~al.
\newblock Improved protein structure prediction using potentials from deep
  learning.
\newblock \emph{Nature}, 577\penalty0 (7792):\penalty0 706--710,
  2020{\natexlab{a}}.

\bibitem[Senior et~al.(2020{\natexlab{b}})Senior, Evans, Jumper, Kirkpatrick,
  Sifre, Green, Qin, {\v{Z}}{\'\i}dek, Nelson, Bridgland,
  et~al.]{senior2020protein}
Senior, A.~W., Evans, R., Jumper, J., Kirkpatrick, J., Sifre, L., Green, T.,
  Qin, C., {\v{Z}}{\'\i}dek, A., Nelson, A.~W., Bridgland, A., et~al.
\newblock Improved protein structure prediction using potentials from deep
  learning.
\newblock \emph{Nature}, 577\penalty0 (7792):\penalty0 706--710,
  2020{\natexlab{b}}.

\bibitem[Shi et~al.(2020{\natexlab{a}})Shi, Xu, Guo, Zhang, and
  Tang]{shi2020graph}
Shi, C., Xu, M., Guo, H., Zhang, M., and Tang, J.
\newblock A graph to graphs framework for retrosynthesis prediction.
\newblock In \emph{International Conference on Machine Learning}, pp.\
  8818--8827. PMLR, 2020{\natexlab{a}}.

\bibitem[Shi et~al.(2020{\natexlab{b}})Shi, Xu, Zhu, Zhang, Zhang, and
  Tang]{shi2020graphaf}
Shi, C., Xu, M., Zhu, Z., Zhang, W., Zhang, M., and Tang, J.
\newblock Graphaf: a flow-based autoregressive model for molecular graph
  generation.
\newblock \emph{arXiv preprint arXiv:2001.09382}, 2020{\natexlab{b}}.

\bibitem[Simm et~al.(2020{\natexlab{a}})Simm, Pinsler, and
  Hern{\'a}ndez-Lobato]{simm2020reinforcement}
Simm, G., Pinsler, R., and Hern{\'a}ndez-Lobato, J.~M.
\newblock Reinforcement learning for molecular design guided by quantum
  mechanics.
\newblock In \emph{International Conference on Machine Learning}, pp.\
  8959--8969. PMLR, 2020{\natexlab{a}}.

\bibitem[Simm \& Hern{\'a}ndez-Lobato(2020)Simm and
  Hern{\'a}ndez-Lobato]{simm2019generative}
Simm, G.~N. and Hern{\'a}ndez-Lobato, J.~M.
\newblock A generative model for molecular distance geometry.
\newblock In \emph{International Conference on Machine Learning}, 2020.

\bibitem[Simm et~al.(2020{\natexlab{b}})Simm, Pinsler, Cs{\'a}nyi, and
  Hern{\'a}ndez-Lobato]{simm2020symmetry}
Simm, G.~N., Pinsler, R., Cs{\'a}nyi, G., and Hern{\'a}ndez-Lobato, J.~M.
\newblock Symmetry-aware actor-critic for 3d molecular design.
\newblock \emph{arXiv preprint arXiv:2011.12747}, 2020{\natexlab{b}}.

\bibitem[Smith et~al.(2020)Smith, Burns, Simmonett, Parrish, Schieber,
  Galvelis, Kraus, Kruse, Di~Remigio, Alenaizan, et~al.]{smith2020psi4}
Smith, D.~G., Burns, L.~A., Simmonett, A.~C., Parrish, R.~M., Schieber, M.~C.,
  Galvelis, R., Kraus, P., Kruse, H., Di~Remigio, R., Alenaizan, A., et~al.
\newblock Psi4 1.4: Open-source software for high-throughput quantum chemistry.
\newblock \emph{The Journal of chemical physics}, 152\penalty0 (18):\penalty0
  184108, 2020.

\bibitem[Smith et~al.(2017)Smith, Isayev, and Roitberg]{smith2017ani}
Smith, J.~S., Isayev, O., and Roitberg, A.~E.
\newblock Ani-1: an extensible neural network potential with dft accuracy at
  force field computational cost.
\newblock \emph{Chemical science}, 8\penalty0 (4):\penalty0 3192--3203, 2017.

\bibitem[Wang et~al.(2020)Wang, Yang, Harris, Gomez-Bombarelli, and
  G{\'{o}}mez-Bombarelli]{Wang2020SIL}
Wang, W., Yang, T., Harris, W.~H., Gomez-Bombarelli, R., and
  G{\'{o}}mez-Bombarelli, R.
\newblock {Active learning and neural network potentials accelerate molecular
  screening of ether-based solvate ionic liquids}.
\newblock \emph{Chemical Communications}, 56\penalty0 (63):\penalty0 8920, aug
  2020.
\newblock ISSN 1359-7345.
\newblock URL
  \url{https://pubs.rsc.org/en/content/articlehtml/2020/cc/d0cc03512b
  https://pubs.rsc.org/en/content/articlelanding/2020/cc/d0cc03512b
  http://pubs.rsc.org/en/Content/ArticleLanding/2020/CC/D0CC03512B}.

\bibitem[Xu et~al.(2018)Xu, Hu, Leskovec, and Jegelka]{xu2018powerful}
Xu, K., Hu, W., Leskovec, J., and Jegelka, S.
\newblock How powerful are graph neural networks?
\newblock \emph{arXiv preprint arXiv:1810.00826}, 2018.

\bibitem[Xu et~al.(2021)Xu, Luo, Bengio, Peng, and Tang]{xu2021learning}
Xu, M., Luo, S., Bengio, Y., Peng, J., and Tang, J.
\newblock Learning neural generative dynamics for molecular conformation
  generation.
\newblock In \emph{International Conference on Learning Representations}, 2021.
\newblock URL \url{https://openreview.net/forum?id=pAbm1qfheGk}.

\bibitem[You et~al.(2018)You, Liu, Ying, Pande, and Leskovec]{you2018graph}
You, J., Liu, B., Ying, Z., Pande, V., and Leskovec, J.
\newblock Graph convolutional policy network for goal-directed molecular graph
  generation.
\newblock In \emph{Advances in neural information processing systems}, pp.\
  6410--6421, 2018.

\end{thebibliography}
